\documentclass[letterpaper]{article}
\usepackage{textcomp}
\DeclareUnicodeCharacter{2013}{--} 
\usepackage[utf8]{inputenc}

\usepackage[margin=0.5in]{geometry} 
\usepackage{array}                
\usepackage{graphicx}             
\usepackage{booktabs}             
\usepackage{arydshln}             
\usepackage{adjustbox}            
\usepackage{rotating}             
\usepackage{makecell}
\usepackage{multirow}

\usepackage{times,amsmath,epsfig}
\usepackage{fancyhdr}
\usepackage{amsmath}
\usepackage{amsfonts}
\usepackage{amssymb}
\usepackage{array}
\usepackage{graphicx}
\usepackage{url}
\usepackage{subfigure}
\usepackage{bm}
\usepackage{breqn}
\usepackage{xcolor}
\usepackage{soul}
\usepackage{amssymb}
\usepackage[breaklinks=true]{hyperref}
\usepackage{breakcites}
\usepackage{microtype}

\begin{document}

\vspace{1 cm}

\title{Evaluating OCR Performance for Assistive Technology: Effects of Walking Speed, Camera Placement, and Camera Type
}

%
\author{ Junchi Feng$^{a,b}$, 
Nikhil Ballem$^{c}$,  
Mahya Beheshti$^{c,d}$, 
Giles Hamilton-Fletcher$^{c}$, 
Todd Hudson$^{c}$, 
\\
Maurizio Porfiri$^{a,b,d,e}$,
William H. Seiple$^{f}$, 
John-Ross Rizzo$^{a,c}$
}
\date{} 
\maketitle

\textbf{Affiliations:}
\begin{itemize}
    \renewcommand{\labelitemi}{}
    \item $^a$ Department of Biomedical Engineering, Tandon School of Engineering, New York University, Brooklyn, NY 11201, USA
    \item $^b$ Center for Urban Science and Progress, Tandon School of Engineering, New York University, Brooklyn, NY 11201, USA
    \item $^c$ Department of Rehabilitation Medicine, NYU Grossman School of Medicine, New York, NY 10016, USA
    \item $^d$ Department of Mechanical and Aerospace Engineering, Tandon School of Engineering, New York University, Brooklyn, NY 11201, USA
    \item $^e$ Department of Civil and Urban Engineering, Tandon School of Engineering, New York University, Brooklyn, NY 11201, USA
    \item $^f$ Lighthouse Guild, New York, NY 10023, USA
\end{itemize}

\clearpage

\begin{abstract}

Optical character recognition (OCR), a process that converts printed or handwritten text into machine-readable form, is widely used in assistive technology for people with blindness and low vision. Yet most evaluations rely on static datasets that do not reflect the challenges of mobile use. In this study, we systematically evaluated OCR performance under both static and dynamic conditions. Static tests measured detection range across distances of 1–7 meters and viewing angles of 0°–75° horizontally. Dynamic tests examined the impact of motion by varying walking speed from slow (0.8 m/s) to very fast (1.8 m/s) and compared three camera mounting positions: head-mounted, shoulder-mounted, and handheld. We evaluated both a smartphone and smart glasses, using the phone’s main and ultra-wide cameras. Four OCR engines were benchmarked to assess accuracy at different distances and viewing angles: Google Vision, PaddleOCR 3.0, EasyOCR, and Tesseract. PaddleOCR 3.0 was then used to evaluate OCR performance under dynamic walking conditions. Accuracy was computed at the character-level using the Levenshtein ratio against manually defined ground truth. Results showed that recognition accuracy declined with increased walking speed and wider viewing angles. Google Vision achieved the highest overall accuracy, with PaddleOCR close behind as the strongest open-source alternative. Across devices, the phone’s main camera achieved the highest accuracy, and a shoulder-mounted placement yielded the highest average among body positions; however, differences among shoulder, head, and hand were not statistically significant.
\end{abstract}

\textbf{Keywords: } Assistive Technology, Computer Vision, Navigation, Optical Character Recognition, Visual Impairment

\section{Introduction}

Visual impairment affects over 300 million people worldwide and is associated with reduced independence, increased reliance on others, higher unemployment, and diminished quality of life \cite{visionAtlas,sherrod2014association,popescu2011age,mckean2007severity,court2014visual}. A major contributor to these challenges is difficulty navigating everyday environments safely and efficiently. Orientation and mobility (O\&M) tasks such as locating entrances, identifying street names, interpreting warning signs, or reading posted instructions often depend on access to environmental text. For people with blindness or low vision (pBLV), this information may be inaccessible or difficult to find, creating barriers to independent travel and participation in daily activities. O\&M research has long emphasized that access to environmental information is critical for safe and confident navigation, and that assistive technologies can complement traditional mobility tools such as long canes or guide dogs by providing additional sources of situational awareness \cite{giudice2008blind,dakopoulos2009wearable}. Assistive technologies therefore play an important role in mitigating these barriers by providing alternative ways to perceive and interpret environmental information \cite{scherer2005assessing, mahmoudi2025transformative}.

One technology that has shown particular promise in this context is optical character recognition (OCR). OCR converts images of printed or handwritten text into machine-readable form, enabling digital systems to detect and interpret textual information from the environment. For pBLV, OCR enables a wide range of assistive applications, including reading printed documents, identifying currency, and recognizing navigation signage in public environments \cite{afb-ocr}. Prior work has explored several OCR-based assistive systems to improve accessibility, ranging from mobile document readers \cite{pundlik2019mobile} to navigation-oriented systems that detect street or facility signage \cite{feng2023commute, zhang2025enhancing}. For instance, mobile applications have been developed that allow users to search for specific keywords in real-world scenes using smartphone cameras, facilitating efficient information access in cluttered environments \cite{pundlik2019mobile}. In navigation contexts, systems such as Commute Booster integrate OCR with transit data to identify relevant wayfinding signage and provide real-time guidance in complex environments such as subway stations \cite{feng2023commute}. Observational studies of assistive technology use also indicate that pBLV frequently rely on access to environmental text to confirm destinations, identify landmarks, and locate entrances during orientation and mobility tasks \cite{kameswaran2020understanding,chanana2017assistive,basiri2016seamless,yu2025visual}. In navigation scenarios, reliable OCR can therefore improve environmental awareness by enabling users to identify street names, read building labels, interpret warning signs, or locate points of interest during travel. When integrated into real-time assistive systems, OCR thus has the potential to reduce uncertainty during navigation and support safer and more independent mobility.

Despite this progress, most OCR performance evaluations rely on static benchmark datasets such as ICDAR Scene Text \cite{shahab2011icdar}, IIIT 5K-Word \cite{MishraBMVC12}, or COCO-Text \cite{veit2016coco}. These datasets contain large collections of labeled images and have been instrumental in advancing OCR algorithms. However, they largely represent static conditions in which the camera and scene remain stable. While several video-based datasets have been introduced to study text recognition in dynamic environments \cite{wu2024dstext, wu2023icdar, nguyen2014video}, these resources primarily focus on algorithmic performance in prerecorded video and provide limited insight into how human motion and real-world device usage affect OCR reliability. This limitation is particularly important for assistive technology applications. During navigation, OCR systems must operate under continuously changing conditions. Users are typically walking, the camera is moving with the body, and environmental text may appear at different distances, angles, and sizes. Motion blur caused by walking can degrade image quality, and wide field-of-view cameras may capture text that appears small or distorted in the frame. Camera mounting location on the body can also influence image stability and viewing geometry. These factors interact with the hardware and software components of the assistive system and can substantially affect recognition reliability. As a result, a sign that is clearly legible in a static image may become difficult or impossible to recognize when captured during natural movement. Despite the practical importance of these factors, the combined effects of walking speed, camera placement, and camera hardware on OCR performance have not been systematically studied.

To address this gap, we designed a controlled experimental framework to evaluate OCR performance under conditions representative of real-world assistive technology use. Specifically, we examined how recognition accuracy changes with walking speed, camera mounting location, and camera field of view. We quantified OCR accuracy across multiple walking speeds, compared camera mounting positions including head-mounted, shoulder-mounted, and handheld configurations, evaluated detection range and recognition accuracy using the iPhone main and ultra-wide cameras as well as the Meta smart glasses, and benchmarked four widely used OCR engines: Tesseract, Google Vision, PaddleOCR, and EasyOCR. We hypothesized that OCR performance would decline with increasing walking speed, wider viewing angles, and wider camera fields of view, and that camera mounting position would influence recognition accuracy through differences in image stability. Because Google Vision is trained on large-scale proprietary datasets, we expected it to achieve the highest overall accuracy. Among open-source engines, we hypothesized that PaddleOCR would provide the strongest performance based on prior evaluations \cite{cui2025paddleocr, feng2025robust,reddy2024license}. By systematically quantifying how mobility-related factors influence OCR accuracy, this study aims to provide empirical guidance for the design of future OCR-based assistive technologies that support safe and effective navigation for people with visual impairments.

\section{Methods}

To evaluate OCR performance under conditions relevant to real-world assistive technology use, we designed a controlled experimental framework combining standardized signage, modern camera-equipped devices, and both static and dynamic testing paradigms. The signage was selected to capture variations in font size, color, and background contrast commonly encountered in public environments. Recordings were collected using consumer devices, including a smartphone and smart glasses, mounted at positions representative of typical wearable configurations. Static tests systematically varied viewing distance and angle to establish recognition range, while dynamic tests examined the impact of user motion and mounting location on recognition stability. All captured frames were then processed by four widely used OCR engines, enabling direct comparison across hardware and software conditions.

\subsection{Signage Setup}

\begin{figure}[ht]
\centering
\includegraphics[width=0.5\textwidth]{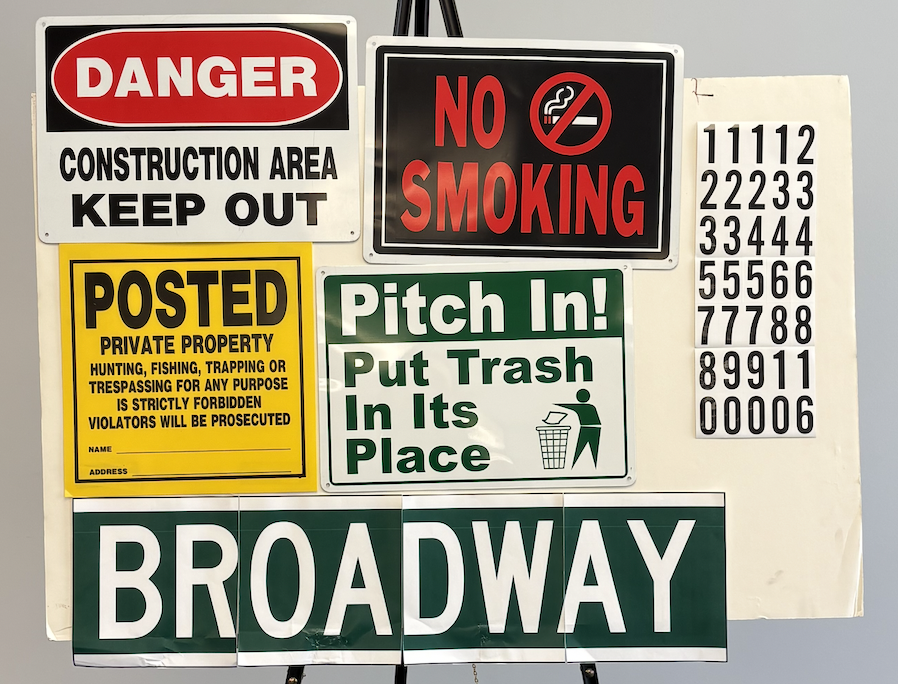}
\caption{
A photo of signage used in the study, illustrating variation in font size, color, and background contrast. 
}
\label{fig:sign}
\end{figure}

As shown in Figure \ref{fig:sign}, we included signage with varied font sizes, colors, and background contrasts to reflect a range of real-world conditions. With the exception of the Broadway sign, all signs were purchased from Home Depot. The ``Danger” sign is a 10 in.~×~14 in. aluminum sign with red text on a black background and black text on a white background \cite{homedepot-construction}. The ``No Smoking” sign is a 10 in.~×~14 in. aluminum sign with red text on a black background \cite{homedepot-nosmoking}. The ``Pitch In!” sign is also 10 in.~×~14 in. aluminum, containing both green-on-white and white-on-green text \cite{homedepot-pitchin}. The “Posted: Private Property” sign is an 11 in.~×~11 in. plastic sign featuring black text on a yellow background in multiple font sizes \cite{homedepot-posted}. The Broadway sign was custom-printed to represent a common New York City street sign, with each character measuring 5 in. in height. Finally, the numeric panel was created using a 2 in. self-adhesive vinyl number set \cite{homedepot-numbers}. These signs were mounted on a board measuring approximately 92 × 70 cm, hereafter referred to as the sign board. Together, these signs represent examples of text commonly encountered in public environments.

\subsection{Hardware Configuration}
\subsubsection{Devices}
\paragraph{Mobile Phone Camera}
\label{sec:static-test}
We used the iPhone 15 Pro Max as the primary input device. This smartphone features a triple rear-camera system representative of current high-end mobile devices. The main camera has a 48 MP sensor with an f/1.78 aperture and an estimated ~84° diagonal field of view, while the ultra-wide camera has a 12 MP sensor with an f/2.2 aperture, a 120° field of view, and a 13 mm lens equivalent. In addition, the telephoto camera includes a 12 MP sensor with an f/2.8 aperture, a 120 mm lens equivalent, and up to 5× optical zoom (25× digital zoom) \cite{wikipedia-iphone15pro}. These camera specs are comparable to most phone cameras on the market, making this device an appropriate representative device for evaluating OCR performance in contemporary mobile platforms.

For this study, we focused on comparing the performance of the main and ultra-wide cameras, as they best capture the trade-off between recognition accuracy and field of view. All recordings were performed using the default settings: 1× zoom; no manual adjustments to brightness, exposure, or filters; standard video mode. Videos were recorded at a resolution of 3840 × 2160 at 30 fps.

\paragraph{Meta Glasses}

We also tested the performance of Ray-Ban Meta Wayfarer Glasses (Gen 1) (model code: RW4006 601ST3 53-22). This device is equipped with an ultra-wide 8 MP camera capable of video acquisition at 1440~×~1920 px at 30 fps \cite{meta-wayfarer}. The glasses have an estimated diagonal field of view of 99° \cite{raybanmeta-brochure}. The glasses are recognized as one of the leading smart glasses on the market \cite{smartglasseson-best}. Beyond market prominence, their lightweight, hands-free design makes them a practical candidate for assistive technology applications, offering a naturalistic form factor that differs from handheld smartphones and better reflects emerging wearable computing trends.

\paragraph{Camera Positions}

\begin{figure}[ht]
\centering
\includegraphics[width=0.6\textwidth]{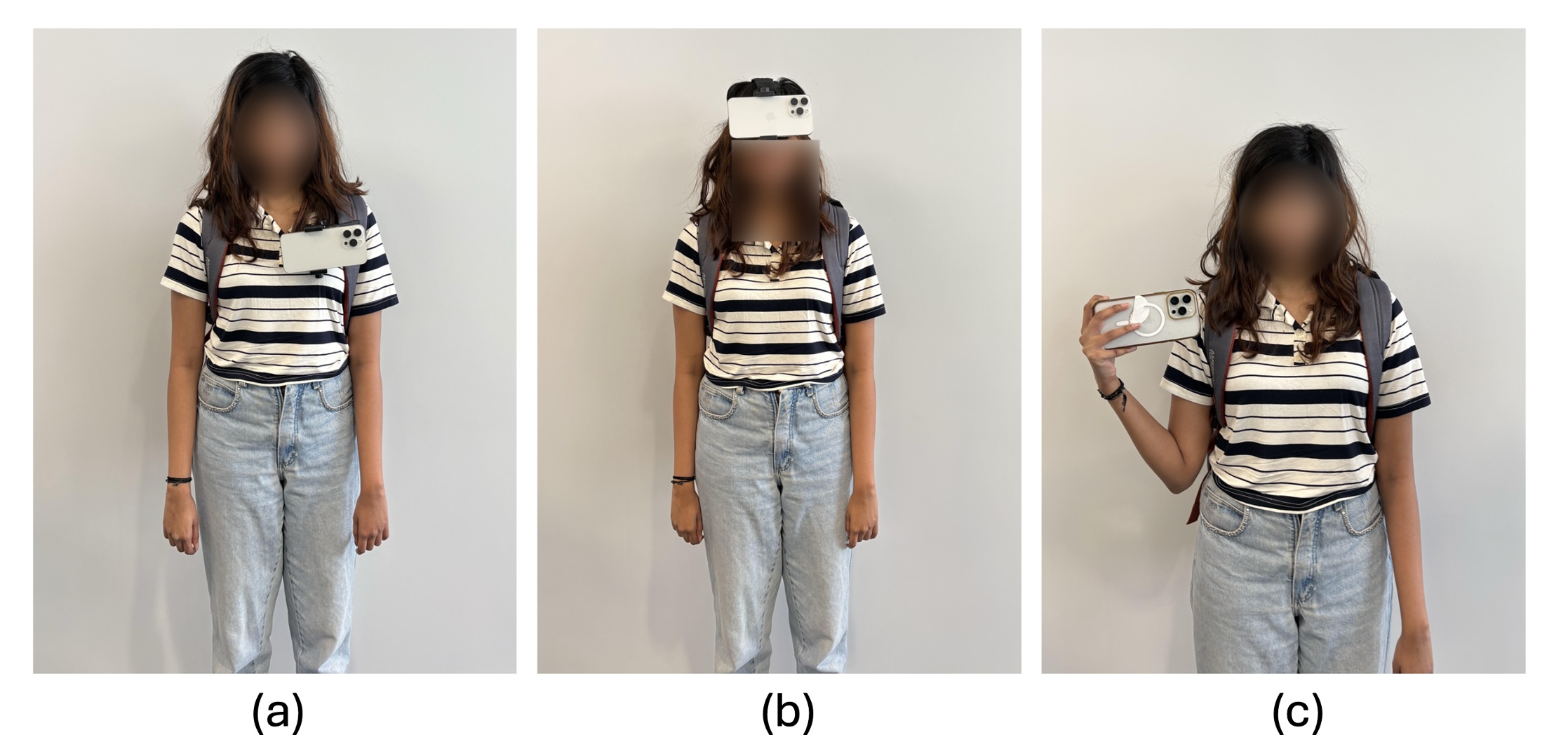}
\caption{
Camera mounting positions evaluated in the study: (a) shoulder-mounted, with the smartphone attached to a backpack strap; (b) head-mounted, with the smartphone secured to a headband; and (c) handheld, with the smartphone held naturally in front of the body.
}

\label{fig:camera_position}
\end{figure}

The Meta glasses were worn in their natural configuration, similar to how eyeglasses are typically used, thereby providing a fixed and realistic camera perspective. In contrast, smartphone placement can vary considerably.  As shown in Figure \ref{fig:camera_position}, we tested three camera positions: shoulder-mounted, head-mounted, and handheld.

In the shoulder-mounted setup, the phone was attached to a backpack shoulder strap at chest-level using a rigid strap mount \cite{rugvis_mount_amazon}. The mount was secured tightly to the shoulder strap to minimize movement. During all walking trials, the backpack’s chest strap was fastened to further stabilize the setup and prevent excessive swinging.

In the head-mounted setup, the phone was attached to an adjustable elastic headband \cite{pellking2025headmount} positioned above the eyes and centered on the forehead. This configuration kept the rear camera forward-facing while avoiding obstruction of the wearer’s vision.

In the handheld setup, the phone was held naturally in the right hand in front of the body at approximately chest-level. In all three positions, the phone was held in landscape orientation.

The volunteer who wore the device and collected the data was 171 cm tall, which placed the camera at approximately 143 cm for the shoulder-mounted position, 166 cm for the head-mounted position, and 139 cm for the handheld position. The volunteer was a healthy adult with normal vision and no known mobility impairments. These positions were selected to represent common categories of camera placement in assistive technologies \cite{han2024wearables}, which are frequently used in sensory substitution devices for pBLV \cite{feng2023commute,ng2022real,hao2022detect,feng2025haptics,ehrlich2017head}.

\subsection{Static Test Setup}

\begin{figure}[ht]
  \centering
  \includegraphics[width=0.6\textwidth]{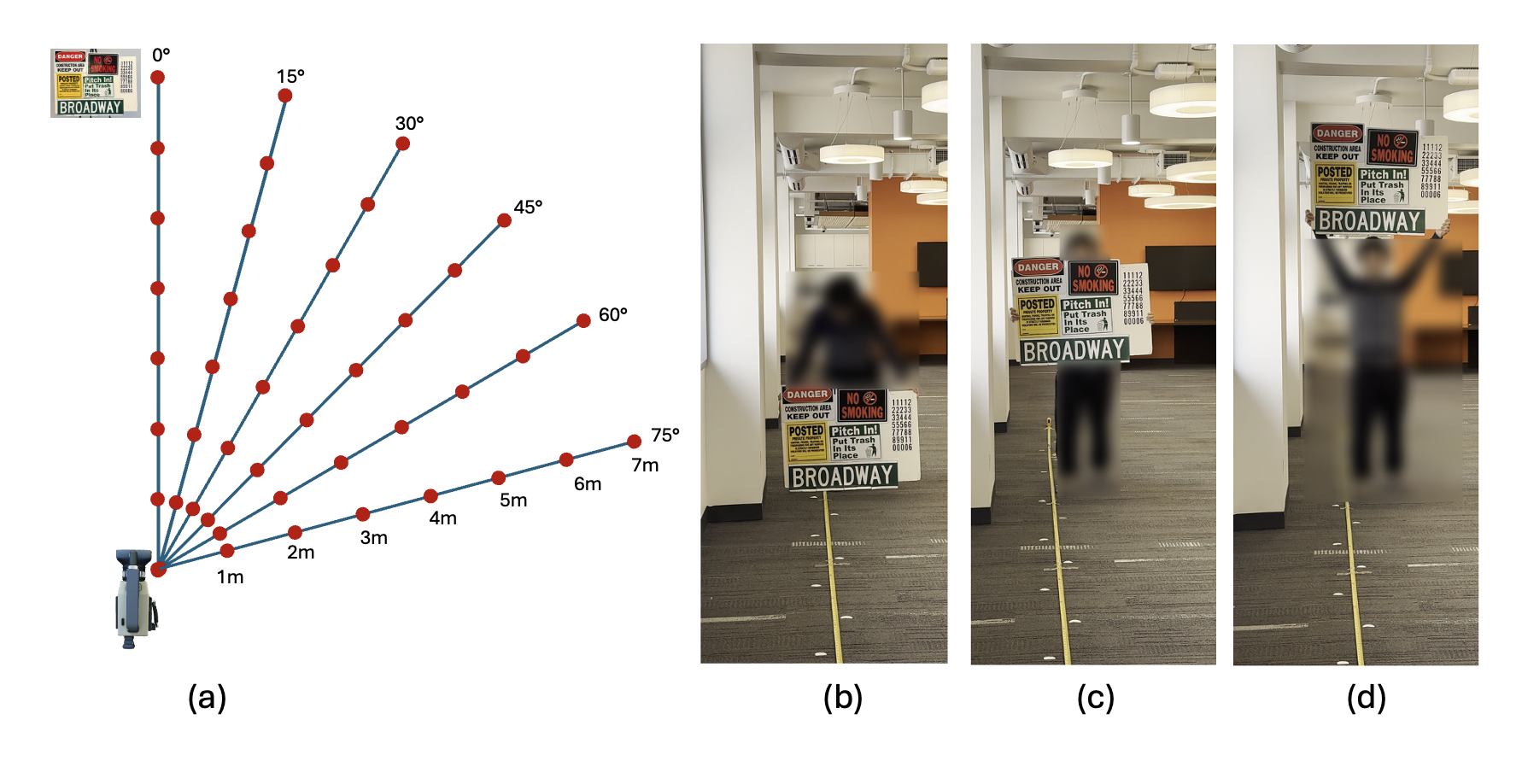}
  \caption{Static test setup for evaluating OCR detection range. (a) Diagram of the test field showing camera placement and sign-board positions at varying distances and angles (red dots); (b) sign board at low height (33 cm); (c) sign board at medium height (100 cm); and (d) sign board at high height (218 cm).}
  \label{fig:static_test_setup}
\end{figure}

This test was designed to determine the detection range of the OCR engines. All experiments were conducted indoors under uniform lighting conditions. The participant remained stationary with the camera in the shoulder-mounted position. The sign board was first placed directly in front of the participant at a distance of 1 meter, then incrementally moved farther away in 1 meter steps up to 7 meters. The same procedure was repeated with the board positioned at 15° to the right of the participant, then at 30°, and continuing in 15° increments. In all conditions, the sign board faced straight ahead, like a fixed street sign: its surface remained perpendicular to the floor and parallel to the participant’s frontal plane, without any tilting or rotation toward the participant. Thus, only the participant’s viewing angle changed, while the sign itself stayed in a constant forward-facing orientation. For the Meta glasses and the iPhone’s main camera, the maximum angle tested was 30°, beyond which the sign was no longer within the field of view. With the iPhone’s ultra-wide camera, the maximum testable angle was 75°. This setup is illustrated in Figure \ref{fig:static_test_setup} (a).

At each distance–angle location (red dots in Figure \ref{fig:static_test_setup}), the board was presented at three heights: low (33 cm), medium (100 cm), and high (218 cm) from the ground to the board center. This setup is shown in Figure \ref{fig:static_test_setup} (b, c, d). Videos were recorded using the settings described in \autoref{sec:static-test}. For each condition, a key frame was extracted from the recording, defined as the first frame in which the sign appeared stable and free of motion blur. These extracted frames, preserved at the original resolution, were then processed by the OCR engines for recognition.

\subsection{Dynamic Test Setup}

This test was designed to examine the impact of walking speed and camera mounting position on OCR detection accuracy. The experiment was conducted in the same environment as the static test. The volunteer carried the camera and walked along a 7-meter straight path at four different speeds: slow (0.8 m/s), medium (1.1 m/s), fast (1.5 m/s), and very fast (1.8 m/s). These speeds represent typical adult walking ranges \cite{bohannon2011normal,alves2020walkability}. To standardize stride length, the volunteer’s preferred stride was measured, and footprints were marked along the path. A metronome was used to control walking pace by cueing steps at specific beats per minute (BPM). The volunteer advanced only when hearing each beat, ensuring nearly constant speed across trials. Training was provided until the target speeds could be maintained consistently. During the experiment, another volunteer monitored walking speed in real time, and any trial deviating by more than ±0.1 m/s from the target was repeated. Speed was measured by timing the 7-meter path with a stopwatch and calculating speed as distance divided by time. Trials were repeated until one valid trial per condition was obtained, defined as a recording that met the ±0.1 m/s speed tolerance and had no recording interruptions.

\begin{figure}[ht]
\centering
\includegraphics[width=0.3\textwidth]{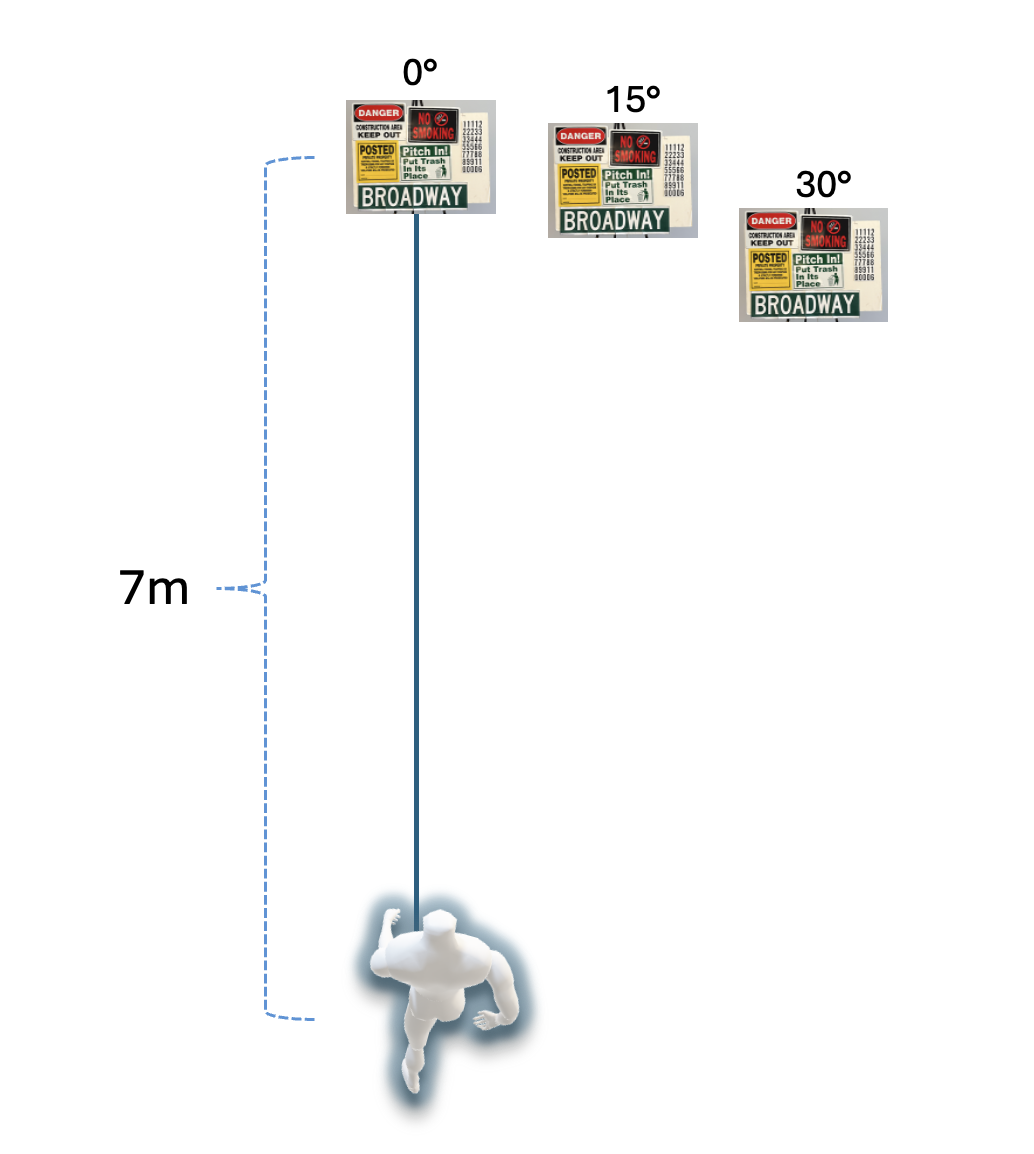}
\caption{
Dynamic test setup. A 7 m straight walkway was marked between a start line and a finish line. The participant walked along this line at four controlled speeds. The sign board was placed so that the line from the start point to the sign subtended an azimuth angle of 0°, 15°, or 30° relative to the walking direction, while the participant’s path remained straight. Thus, at 0° the participant walked directly toward the sign, and at 15° and 30° the sign was laterally offset from the path.
}
\label{fig:dynamic_testing_setup}
\end{figure}

The sign board was mounted on a tripod, oriented nearly perpendicular to the floor, with its center positioned 120 cm above ground-level. For each trial, the board was placed at 7 meters directly in front of the volunteer (0°), then repositioned at 15°, and finally at 30° relative to the walking path.  For each speed and angle, all three camera positions were tested. This setup is illustrated in Figure \ref{fig:dynamic_testing_setup}.

\subsection{OCR Models}

We evaluated four OCR engines to assess performance: PaddleOCR, Tesseract, EasyOCR, and Google Vision.

\textbf{PaddleOCR.} We employed PaddleOCR 3.0, an open-source OCR toolkit developed by Baidu on the PaddlePaddle deep learning framework \cite{paddleocr}. The developers highlight its ultra-lightweight design, high accuracy, and fast inference speed \cite{paddleocr}, which has drawn significant attention since its release. Independent evaluations further confirm its superior performance compared to other open-source OCR systems \cite{hadi2024benchmarking}. These factors make PaddleOCR a strong candidate for assistive technology applications, motivating its inclusion in our study.

\textbf{Tesseract.} Tesseract is one of the earliest open-source OCR engines, developed in the 1980s and released as open source in 2005 \cite{smith2007overview,smith2013history}. Now maintained by Google, it supports over 100 languages, is actively updated, and is easily integrated into diverse applications \cite{tesseract2025}. Owing to its maturity, open-source availability, and extensive use in both academic research and commercial applications, Tesseract serves as a natural baseline for benchmarking OCR performance.  Tesseract-ocr version 4.1.1 was used for this project.

\textbf{EasyOCR.} EasyOCR is an open-source OCR library developed by the AI company Jaided AI \cite{easyocr2025}. It supports more than 80 languages and provides pretrained deep learning models that can be used with minimal setup. EasyOCR is designed for simplicity and rapid prototyping, requiring only a few lines of code to process text in images. Its lightweight architecture and developer-friendly design have made it widely adopted in accessibility and research applications. We included EasyOCR to represent a fast, community-driven alternative that prioritizes ease of integration while still offering competitive recognition accuracy. EasyOCR version 1.7.2 was used for this project.

\textbf{Google Vision.} Google Vision OCR is part of Google Cloud’s vision API, a commercial service that applies large-scale deep learning models trained on diverse real-world datasets \cite{google-vision-ocr}. It is widely recognized for high accuracy in scene text recognition, including complex backgrounds and varied fonts, and it supports multiple languages. As a cloud-based solution, it benefits from continuous updates and integration with Google’s broader AI ecosystem. We included Google Vision OCR to serve as a high-performance commercial benchmark, providing an upper bound for comparison with open-source alternatives. Google Cloud Vision version 3.10.1 was used for this project.

For the static testing, each image and video frame in our dataset was processed by all four models to enable direct comparison. For the dynamic testing, all results were generated using PaddleOCR. This choice was based on the static results (presented in the following section), which showed that PaddleOCR achieved accuracy comparable to Google Vision while remaining free to run locally. In contrast, Google Vision requires paid cloud processing for each image, making it impractical for continuous video data. The other two OCR engines performed too poorly to provide meaningful comparisons and were therefore excluded from dynamic testing.

\subsection{Data Analysis}
We entered the exact ground truth text for each sign to match the printed content. Because the same board and layout were used in all trials, the spatial arrangement of the signs was constant.

Each OCR engine outputs quadrilateral bounding boxes around detected text. To align these detections with the known board layout, we used the \textit{DANGER} sign at the top left corner of the board as a reference anchor. In each frame, the bounding box at the top left was taken as the anchor candidate. We then compared its recognized text against the word \textit{DANGER} using the Levenshtein ratio. If the similarity score exceeded 0.5, the anchor was automatically confirmed. Otherwise, the detection for the \textit{DANGER} sign might be incorrect, so we manually verified whether the box actually covered the \textit{DANGER} sign. If it did, we still used that box as the geometric anchor. If no anchor was detected, the sign board was considered out of view and the frame was skipped.

Once the anchor was fixed, the locations of the other signs were inferred from their constant relative positions on the board. For example, the sign immediately to the right of the \textit{DANGER} anchor was always the \textit{NO SMOKING} sign. Thus, the bounding box detected in that position was assigned to the \textit{NO SMOKING} label. The same procedure was applied to all remaining signs based on their relative layout. All assignments were manually checked to ensure accuracy.

For each text line within each visible sign, we calculated character-level recognition accuracy using the Levenshtein ratio \cite{Levenshtein1966, yujian2007normalized}:
\[
\text{Accuracy}^{(\ell)} \;=\; \frac{\lvert s_{\mathrm{OCR}}^{(\ell)} \rvert + \lvert s_{\mathrm{GT}}^{(\ell)} \rvert - \mathrm{LevDist}\!\left(s_{\mathrm{OCR}}^{(\ell)}, s_{\mathrm{GT}}^{(\ell)}\right)}{\lvert s_{\mathrm{OCR}}^{(\ell)} \rvert + \lvert s_{\mathrm{GT}}^{(\ell)} \rvert}, \quad 0 \le \text{Accuracy}^{(\ell)} \le 1.
\]
Here, for each line $\ell$, $s_{\mathrm{OCR}}^{(\ell)}$ is the OCR string and $s_{\mathrm{GT}}^{(\ell)}$ is the corresponding ground truth string. The terms $\lvert s_{\mathrm{OCR}}^{(\ell)} \rvert$ and $\lvert s_{\mathrm{GT}}^{(\ell)} \rvert$ denote their lengths in characters, and $\mathrm{LevDist}(\cdot,\cdot)$ is the Levenshtein edit distance, defined as the minimum number of single character insertions, deletions, or substitutions required to transform one string into the other.

Frame-level OCR accuracy was defined as the mean of the line-level accuracies across all lines in all visible signs in that frame:
\[
\text{Accuracy}_{\mathrm{frame}} \;=\; \frac{1}{L_f}\,\sum_{\ell=1}^{L_f} \text{Accuracy}^{(\ell)},
\]
where $L_f$ is the total number of evaluated lines in the frame. These values were summarized using heatmaps, line plots, and box plots. Because accuracy scores are bounded in $[0,1]$ and typically non-normal, we used non-parametric tests throughout. Group differences were assessed using the Kruskal--Wallis test, with Dunn's or Mann–Whitney U tests for post hoc comparisons. All $p$ values were adjusted for multiple testing using the Holm procedure \cite{kruskal1952use, holm1979simple,mann1947test}.

We also studied trial-level OCR accuracy for the dynamic tests. Trial-level accuracy was computed as the arithmetic mean of the frame-level accuracies across all frames in that walk:
\[
\text{Accuracy}_{\mathrm{trial}} \;=\; \frac{1}{F}\,\sum_{f=1}^{F} \text{Accuracy}_{\mathrm{frame}}^{(f)},
\]
where $F$ is the number of analyzed frames in the trial. This provides a concise summary of performance for each tested condition.

To obtain an overall comparison of mounting positions in the dynamic test, we performed an additional aggregate analysis using trial-level OCR accuracy. These trial-level values were pooled across all dynamic conditions and grouped by mounting position (handheld, shoulder-mounted, and head-mounted). Overall performance for each mounting position was then defined as the mean trial-level OCR accuracy across all combinations of walking speed, viewing angle, and camera type. Differences among mounting positions were evaluated using a Kruskal--Wallis test, followed by pairwise Mann--Whitney U tests with Bonferroni correction.

Unless otherwise indicated, all accuracy values reported in the text, tables, and figures are proportion correct in the range \([0,1]\).

\section{Results}

\subsection{Static Testing}

\subsubsection{Main Camera}

\begin{figure}[ht]
\centering
\includegraphics[width=1\textwidth]{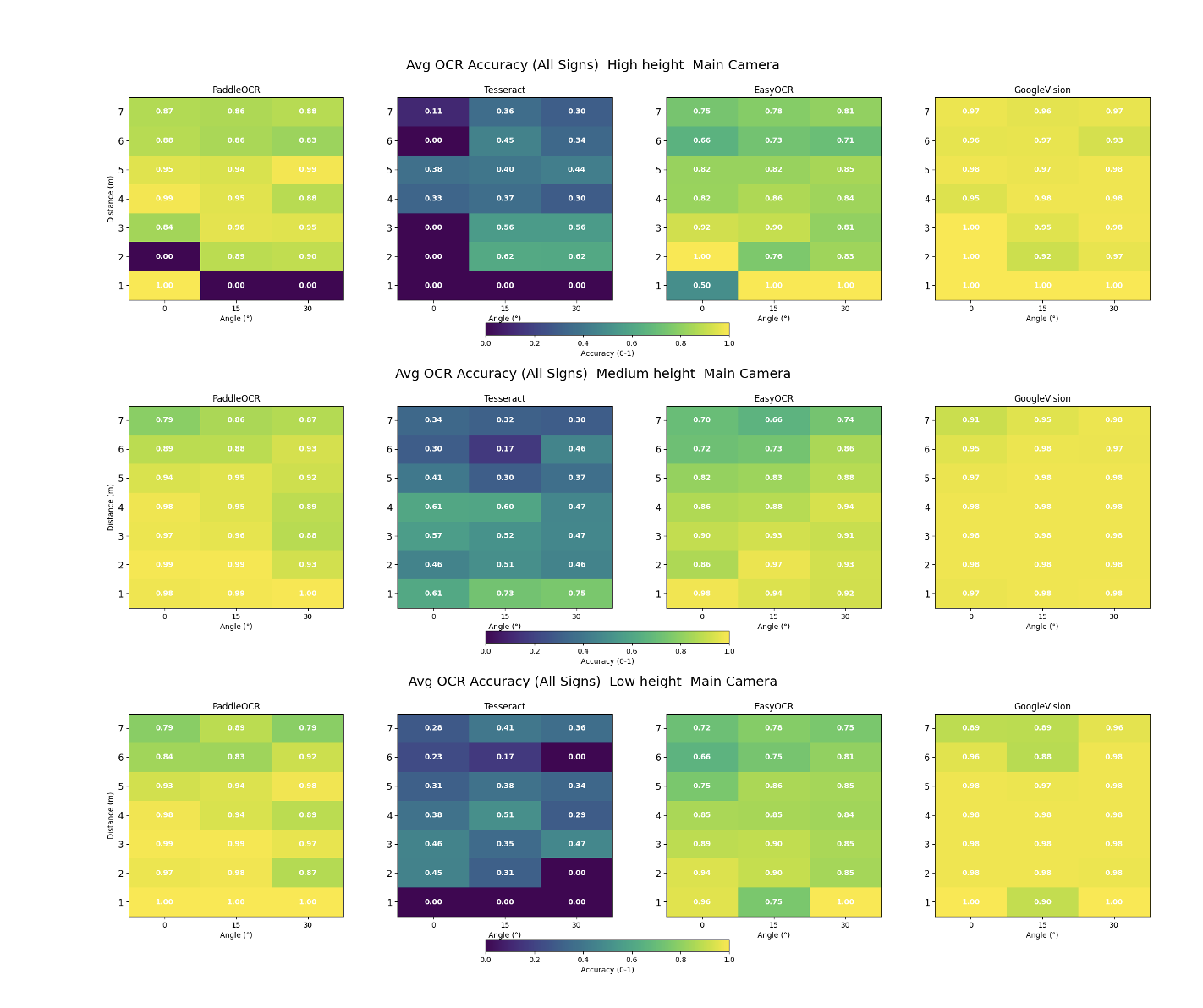}
\caption{
Heatmaps of average OCR accuracy for the main camera at the high, medium, low heights from top to bottom. Each panel shows character-level accuracy for one engine (PaddleOCR, Tesseract, EasyOCR, Google Vision) across distances (1–7 m, y-axis) and viewing angles (0°, 15°, 30°, x-axis). Brighter colors indicate higher accuracy. 
}
\label{fig:static_low}
\end{figure}

To visualize OCR performance across spatial configurations, we generated heatmaps of average character-level accuracy for each engine (Figure \ref{fig:static_low}). Each heatmap displays accuracy values across viewing distances (1–7 m) and angles (0°, 15°, 30°). The main camera consistently showed a clear performance hierarchy: Google Vision and PaddleOCR achieved the highest accuracy across nearly all conditions, EasyOCR yielded intermediate values, and Tesseract was consistently lowest with large patches of near-zero accuracy. These visualizations complement the statistical results by illustrating the robustness of Google Vision and PaddleOCR to changes in distance and angle, whereas Tesseract degrades substantially even under favorable conditions.

For the main camera (all heights pooled), the mean accuracies were: Google Vision (Mean = 0.967, Var = 0.0096), PaddleOCR (Mean = 0.917, Var = 0.0396), EasyOCR (Mean = 0.828, Var = 0.0580), and Tesseract (Mean = 0.380, Var = 0.1127). A Kruskal--Wallis test confirmed significant differences across engines (\(p < 0.001\)). Post-hoc Dunn’s tests indicated that Google Vision was significantly more accurate than both EasyOCR and Tesseract (all adjusted \(p < 0.001\)) but not significantly different from PaddleOCR (adjusted \(p > 0.8\)). PaddleOCR was also significantly more accurate than Tesseract (adjusted \(p < 0.001\)) and EasyOCR (adjusted \(p < 0.001\)), while EasyOCR outperformed Tesseract (adjusted \(p = 0.020\)).

OCR accuracy did not significantly differ across heights for PaddleOCR (\(H = 3.85, p = 0.146\)) or Google Vision (\(H = 0.64, p = 0.728\)), indicating that these engines were robust to vertical placement of the sign board. In contrast, Tesseract showed a strong effect of height (\(H = 41.66, p < 0.001\)), with significantly higher accuracy at the medium height compared to both low (\(p < 0.001, \delta = -0.24\)) and high (\(p < 0.001, \delta = 0.18\)). EasyOCR also showed a weaker but significant effect (\(H = 13.56, p = 0.001\)), with medium height outperforming both low (\(p = 0.011, \delta = -0.10\)) and high (\(p = 0.002, \delta = 0.13\)). Here, \(\delta\) denotes Cliff’s delta effect size, with negative values indicating that the first group tended to have lower scores than the second.

\subsubsection{Wide Camera}

\begin{figure}[ht]
\centering
\includegraphics[width=0.9\textwidth]{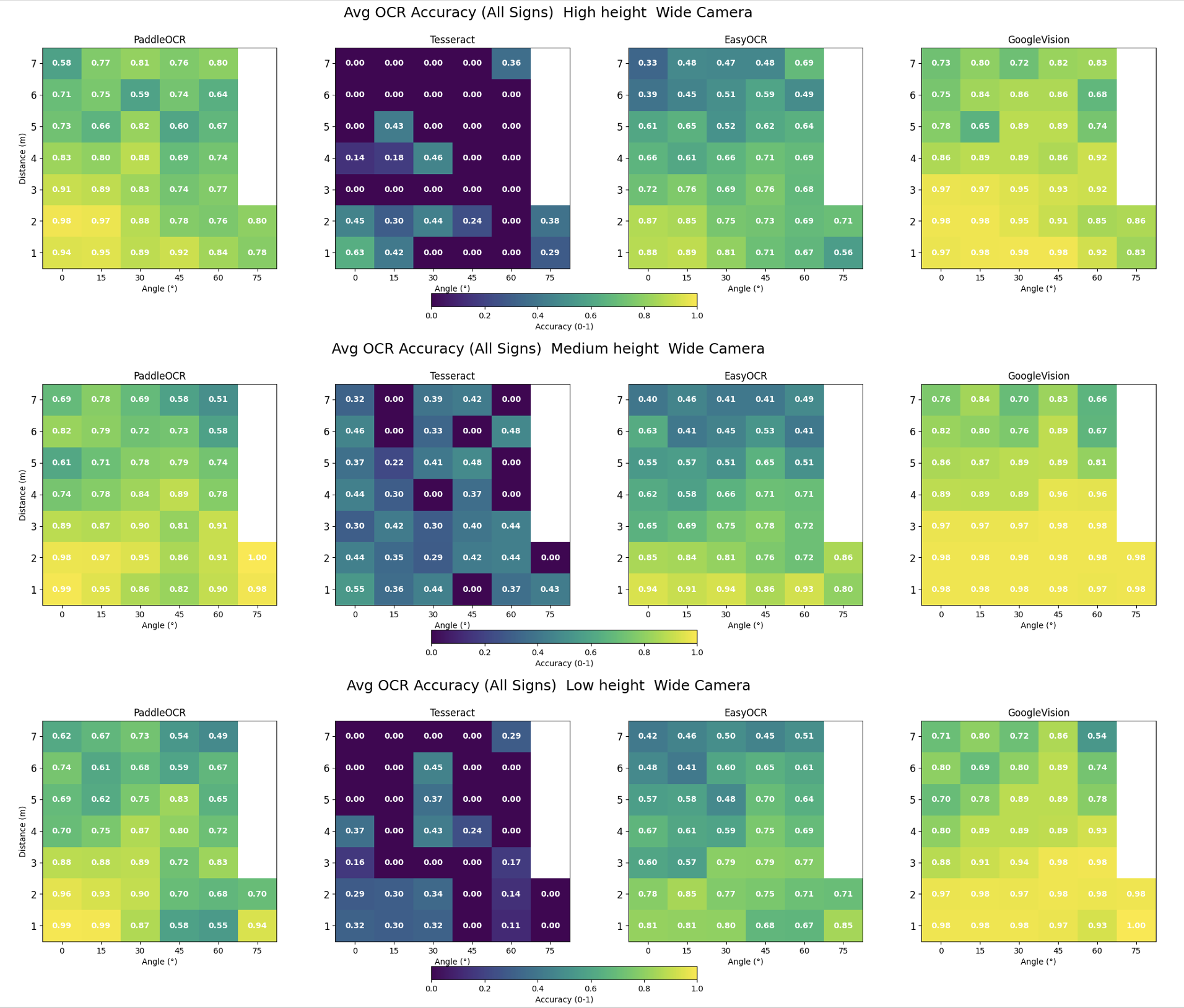}
\caption{
Heatmaps of average OCR accuracy for the ultra-wide camera at the high, medium, low heights from top to bottom. Each panel shows character-level accuracy for one engine (PaddleOCR, Tesseract, EasyOCR, Google Vision) across distances (1–7 m, y-axis) and viewing angles (0°, 15°, 30°, 45°, 60°, 75° x-axis). Brighter colors indicate higher accuracy.
}
\label{fig:static_wide_low}
\end{figure}

Heatmaps for the ultra-wide camera (Figure \ref{fig:static_wide_low}) show a similar ranking of OCR engines but with overall lower accuracy compared to the main camera. Google Vision remains the most accurate across conditions, though its performance decreases more noticeably at longer distances and wide viewing angles. PaddleOCR maintains moderate accuracy but is less stable in the ultra-wide camera compared to the main camera, while EasyOCR shows further reductions in performance. Tesseract again performs consistently poorly, with many conditions yielding near-zero accuracy. These results highlight that the ultra-wide camera introduces additional challenges for OCR performance, although the relative hierarchy among engines is preserved.

For the ultra-wide camera (all heights pooled), the mean accuracies were: Google Vision (Mean = 0.887, Var = 0.0499), PaddleOCR (Mean = 0.788, Var = 0.0941), EasyOCR (Mean = 0.655, Var = 0.0912), and Tesseract (Mean = 0.190, Var = 0.0777). Again, the Kruskal--Wallis test was highly significant (\(p < 0.001\)). Dunn’s tests showed that Google Vision was significantly more accurate than all other engines (all adjusted \(p < 0.001\)). PaddleOCR was significantly more accurate than both EasyOCR and Tesseract (all adjusted \(p < 0.001\)), and EasyOCR also significantly outperformed Tesseract (adjusted \(p < 0.001\)).

For PaddleOCR, accuracy varied significantly across heights (\(H = 12.28, p = 0.002\)), with medium height yielding higher accuracy than low height (\(p = 0.001, \delta = -0.09\)); other contrasts were not significant. Tesseract showed a very strong height effect (\(H = 233.92, p < 0.001\)), with medium height substantially outperforming both low (\(p < 0.001, \delta = -0.34\)) and high (\(p < 0.001, \delta = 0.37\)); low vs. high was also weakly different (\(p = 0.021, \delta = 0.06\)). EasyOCR showed no significant height effect (\(H = 1.61, p = 0.447\)). Google Vision showed a modest effect of height (\(H = 6.76, p = 0.034\)), with medium height significantly outperforming high (\(p = 0.031, \delta = 0.06\)), while other contrasts were not significant. Overall, medium height tended to maximize accuracy for the ultra-wide camera, particularly for Tesseract and to a lesser extent for PaddleOCR and Google Vision, while EasyOCR was unaffected.

\subsubsection{Meta Glasses}

\begin{figure}[ht]
\centering
\includegraphics[width=0.9\textwidth]{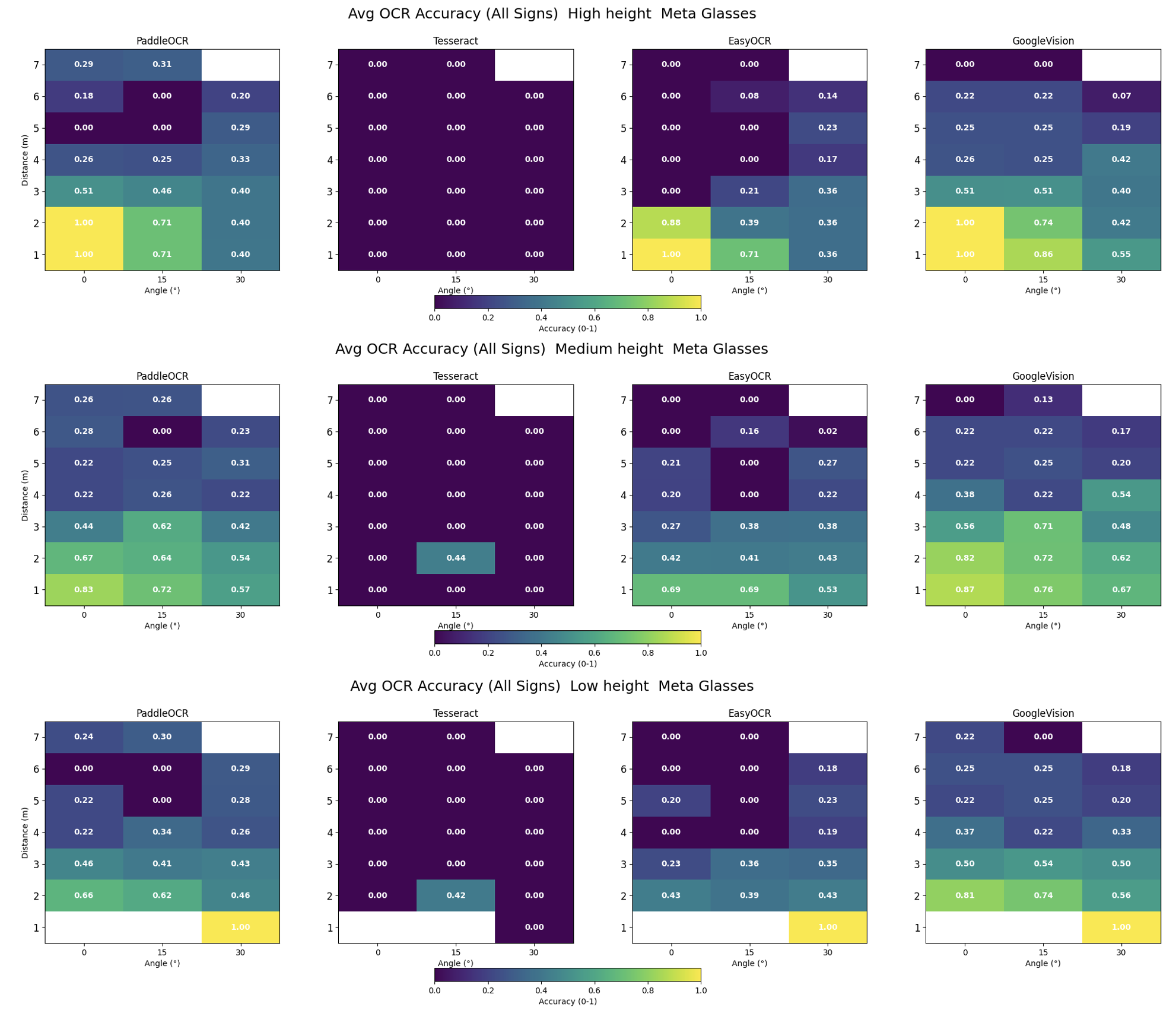}
\caption{Heatmaps of average OCR accuracy for Meta glasses across three mounting heights (high, medium, low) and four OCR engines (PaddleOCR, Tesseract, EasyOCR, Google Vision). Accuracy values (0--1) are averaged over all signs at each tested distance (y-axis, 1--7 m) and camera angle (x-axis, 0\textdegree, 15\textdegree, 30\textdegree).}
\label{fig:meta_glasses_accuracy}
\end{figure}

Figure \ref{fig:meta_glasses_accuracy} is the accuracy heatmap for the Meta glasses. Google Vision again achieved the best results (M = 0.378, Var = 0.111), with PaddleOCR close behind (M = 0.343, Var = 0.083). EasyOCR yielded moderate accuracy (M = 0.205, Var = 0.057), while Tesseract performed poorly (M = 0.018, Var = 0.009).

A Kruskal Wallis test revealed significant differences among the engines (\(H=1184.55,\ df=3,\ p \ll 10^{-3}\)). Post hoc Dunn’s tests showed that Google Vision and PaddleOCR both significantly outperformed EasyOCR and Tesseract (all adjusted \(p \ll 10^{-3}\)), and Google Vision maintained a significant advantage over PaddleOCR (adjusted \(p \ll 10^{-3}\)). In contrast, EasyOCR and Tesseract did not differ significantly (adjusted \(p \approx 1.0\)). P values were adjusted using the Holm method.

Taken together, these results indicate that although Google Vision remains the most accurate engine and PaddleOCR continues to closely follow, the absolute accuracy achieved with Meta glasses is substantially lower than with the iPhone 15 Pro Max.

All four OCR engines showed significant differences across heights when tested with Meta glasses. PaddleOCR exhibited a moderate effect (\(H = 26.27, p < 0.001\)), with medium height outperforming both low (\(p = 0.001, \delta = -0.16\)) and high (\(p < 0.001, \delta = 0.24\)). Tesseract also showed a height effect (\(H = 13.46, p = 0.001\)), driven mainly by higher accuracy at low and medium compared with high (both \(p < 0.001\), \(\delta \approx 0.05{-}0.06\)). EasyOCR showed the strongest dependence on height (\(H = 53.78, p < 0.001\)), with medium height outperforming both low (\(p < 0.001, \delta = -0.18\)) and high (\(p < 0.001, \delta = 0.34\)), and low also exceeding high (\(p = 0.001, \delta = 0.18\)). Google Vision likewise showed a strong effect (\(H = 38.24, p < 0.001\)), with medium height significantly higher than both low (\(p = 0.005, \delta = -0.13\)) and high (\(p < 0.001, \delta = 0.30\)), and low also higher than high (\(p < 0.001, \delta = 0.18\)). Overall, medium height consistently yielded the best OCR performance across engines when using Meta glasses, while high placement was most detrimental.

\subsubsection{Cross-device comparison under static conditions}
To quantify device performance in the static test, we pooled all heights and summarized character-level accuracy for the iPhone main camera, the iPhone ultra-wide camera, and Meta glasses across engines. Table~\ref{tab:static_abs_diffs_named} reports absolute means and absolute gaps between devices.

Across engines, the iPhone main camera achieved the highest absolute accuracy. The iPhone ultra-wide camera was lower than the main camera but remained well above Meta glasses. Meta glasses showed the lowest accuracy.

\begin{table}[ht]
\centering
\caption{Static testing with all heights pooled. Mean character-level accuracy by device and absolute differences. Differences are defined as accuracy (first device) minus accuracy (second device); positive values mean the first device is higher.}
\label{tab:static_abs_diffs_named}
\begin{tabular}{lcccccc}
\toprule
Engine & Mean main & Mean wide & Mean Meta & Difference main to wide & Difference main to Meta & Difference wide to Meta \\
\midrule
Google Vision & 0.967 & 0.887 & 0.378 & 0.080 & 0.589 & 0.509 \\
PaddleOCR     & 0.917 & 0.788 & 0.343 & 0.129 & 0.574 & 0.445 \\
EasyOCR       & 0.828 & 0.655 & 0.205 & 0.173 & 0.623 & 0.450 \\
Tesseract     & 0.380 & 0.190 & 0.018 & 0.190 & 0.362 & 0.172 \\
\bottomrule
\end{tabular}
\end{table}

In summary, absolute gaps highlight clear separations between devices. With Google Vision the main camera exceeded the ultra-wide camera by 0.080 and exceeded Meta glasses by 0.589, while the ultra-wide camera exceeded Meta glasses by 0.509. Similar patterns were observed for the other engines.

\subsection{Dynamic Testing}

\subsubsection{Heatmap}

\begin{figure}[ht]
\centering
\includegraphics[width=1\textwidth]{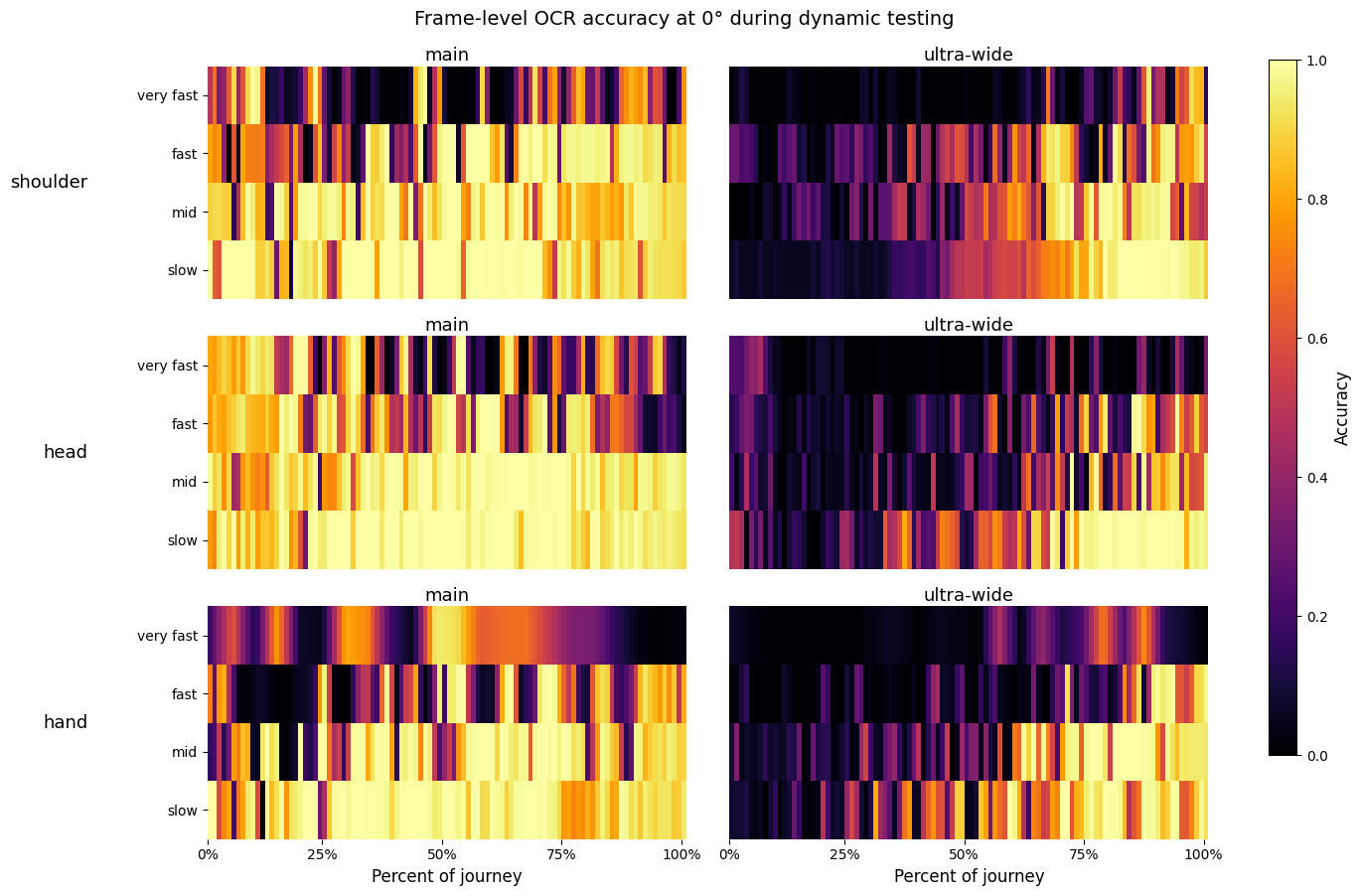}
\caption{
Frame-level OCR accuracy at \(0^{\circ}\) angle during dynamic testing as a function of walking speed (rows) and camera position (shoulder, head, hand) for the main camera (left column) and ultra-wide camera (right column). The x-axis represents the percent of the 7-meter walking trajectory, and the color scale indicates accuracy from low (dark) to high (bright). The main camera generally maintains higher and more stable accuracy, while the ultra-wide camera shows greater variability and lower performance, especially at the start of the approach and under faster walking speeds.
}
\label{fig:dynamic_test_heatmap}
\end{figure}

In the dynamic testing condition, OCR accuracy diminished as walking speed increased (Figure~\ref{fig:dynamic_test_heatmap}). For clarity we present only the \(0^{\circ}\) viewing angle; heatmaps at \(15^{\circ}\) and \(30^{\circ}\) show the same qualitative pattern, so \(0^{\circ}\) serves as a representative example. The decline in accuracy was most evident at the start of the trajectory, approximately 7 meters from the sign, where recognition accuracy was lowest. The ultra-wide camera consistently showed poorer accuracy at the beginning of the approach compared with the main camera, although its performance improved as the camera moved closer to the sign. By contrast, the main camera maintained relatively higher accuracy throughout, though performance still declined at faster walking speeds. These patterns motivated a more detailed analysis of how camera position, camera type, sign angle, and walking speed interact to influence OCR performance, presented in the following section.

\subsubsection{Accuracy vs Speed}

\begin{figure}[ht]
\centering
\includegraphics[width=1\textwidth]{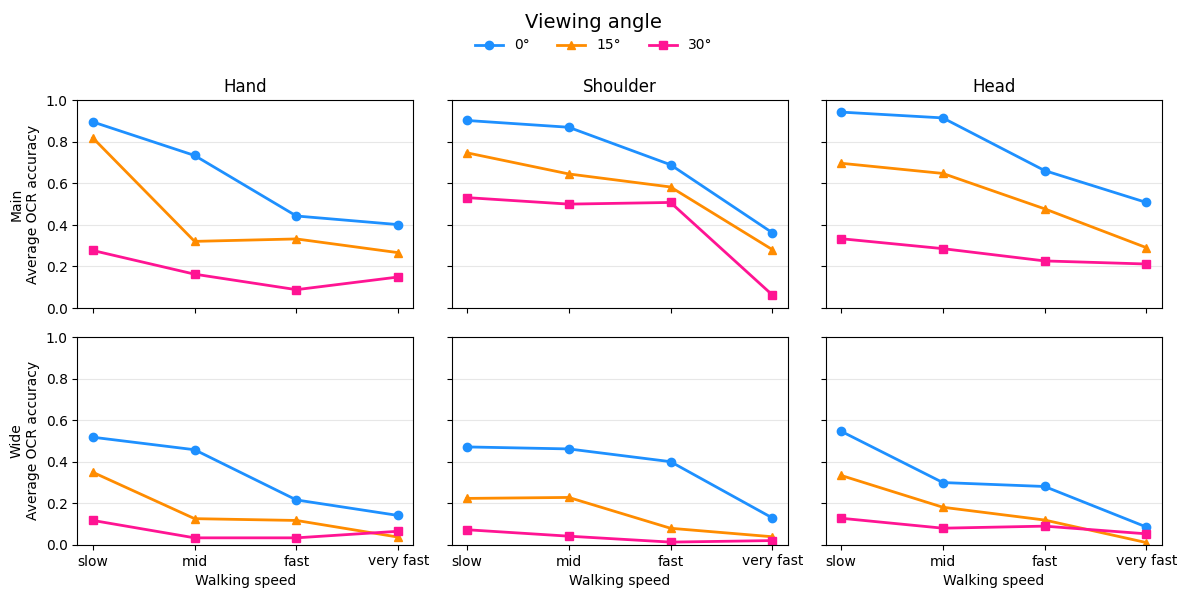}
\caption{Average OCR accuracy across walking speeds for different camera positions (Hand, Shoulder, Head) and camera types (main vs. wide). 
Each subplot corresponds to one camera position and camera type combination, with walking speed on the $x$-axis and OCR accuracy on the $y$-axis. 
Line color and marker shape encode the viewing angle (0° = blue circles, 15°= orange triangles, 30° = green squares). }
\label{fig:ocr_speed}

\end{figure}

Figure~\ref{fig:ocr_speed} shows average trial accuracy across speeds, broken down by body position, camera type, and sign angle. Accuracy declined systematically as speed increased: mean trial accuracy was 0.495 (Var = 0.082) at slow speed, 0.389 (Var = 0.079) at medium speed, 0.298 (Var = 0.049) at fast speed, and 0.173 (Var = 0.022) at very fast speed. A Kruskal--Wallis test confirmed a significant effect of speed (\(H = 14.38, p = 0.0024\)). Post-hoc tests indicated that the most reliable differences occurred between slow vs. very fast and medium vs. very fast conditions (Holm-adjusted \(p < 0.05\)), whereas intermediate contrasts were not significant. Walking speed influenced performance: accuracy was similar at slow, medium, and fast speeds, with a marked decline at very fast walking.

\subsubsection{Accuracy vs Angle}

\begin{figure}[ht]
\centering
\includegraphics[width=1\textwidth]{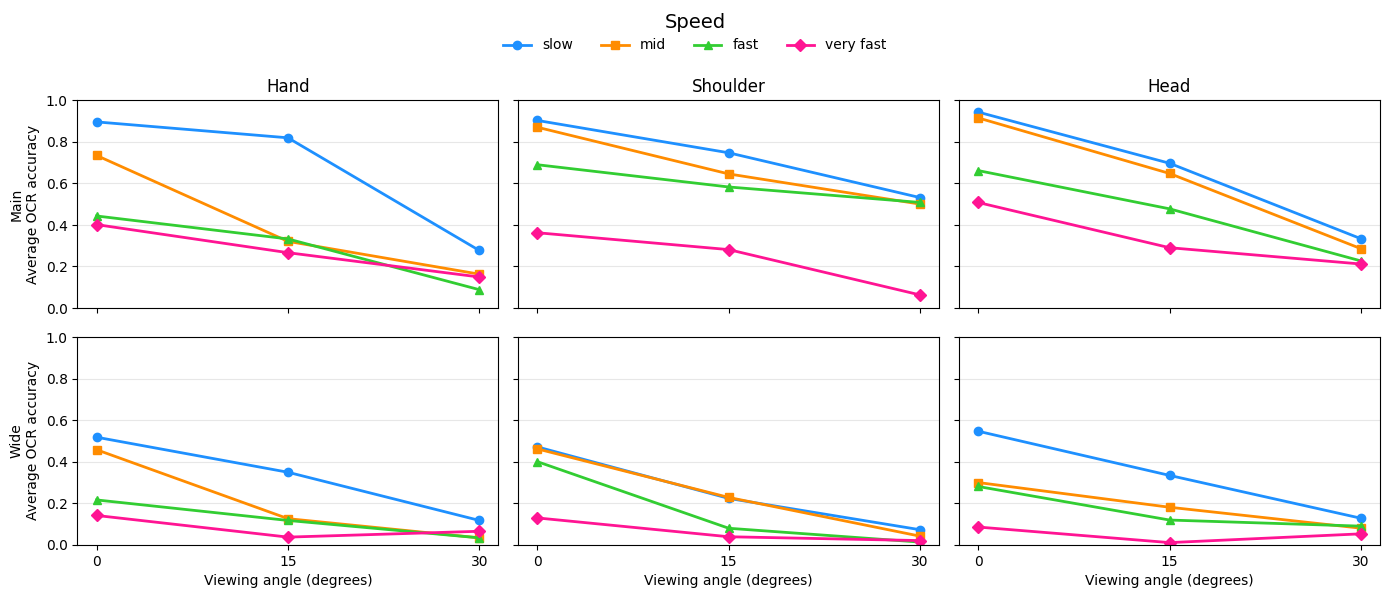}
\caption{Trial-level OCR accuracy by viewing angle for the iPhone. 
Columns show camera position (Hand, Shoulder, Head). Rows show camera type (main on top, ultra-wide on bottom).
Each line represents a walking speed (slow, medium, fast, very fast). The x-axis is viewing angle (0°, 15°, 30°).
Values are trial-level means computed from frame-level accuracies within each condition. }

\label{fig:ocr_angle}
\end{figure}

We next examined how viewing angle influenced trial-level OCR accuracy (Figure~\ref{fig:ocr_angle}). Average accuracy decreased systematically with increasing angle: 0.514 (Var = 0.070) at 0°, 0.331 (Var = 0.059) at 15°, and 0.170 (Var = 0.025) at 30°. A Kruskal--Wallis test confirmed a significant effect of angle (\(H = 21.23, p < 0.001\)). Pairwise Mann–Whitney U tests (Holm-adjusted) showed that all contrasts were significant: recognition accuracy was significantly higher at 0° compared to both 15° and 30° (\(p < 0.05\)), and accuracy at 15° was significantly higher than at 30° (\(p < 0.05\)). These findings demonstrate that OCR performance is strongly degraded by oblique viewing, with a clear monotonic drop as angle increases.

\subsubsection{Accuracy vs Camera Position by Speed}

\begin{figure}[ht]
\centering
\includegraphics[width=1\textwidth]{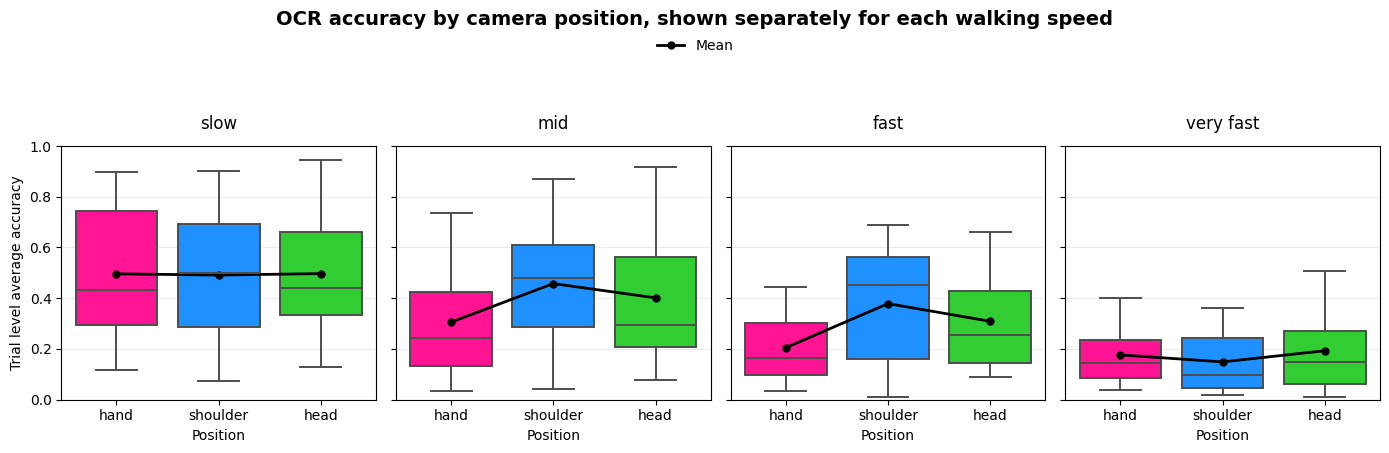}
\caption{
Distribution of trial-level OCR accuracy across camera positions (hand, shoulder, head), shown separately for each walking speed (slow, medium, fast, very fast).
}
\label{fig:ocr_position_speed}
\end{figure}

We evaluated whether mounting position (hand, shoulder, head) influenced trial-level OCR accuracy at each walking speed. The distribution of trial-level accuracy across camera positions and walking speeds is shown in Figure \ref{fig:ocr_position_speed}. At \emph{slow} speed, mean accuracies were nearly identical (hand \(=0.496\), shoulder \(=0.491\), head \(=0.497\)), and no statistical differences were found (\(H=0.04, p=0.983\)). At \emph{medium} speed, the shoulder-mounted camera showed the highest mean accuracy (shoulder \(=0.458\), head \(=0.401\), hand \(=0.306\)), but this effect was not significant (\(H=0.98, p=0.612\)). At \emph{fast} speed, the same pattern was observed (shoulder \(=0.379\), head \(=0.309\), hand \(=0.205\)), yet again without significance (\(H=1.51, p=0.470\)). At \emph{very fast} speed, mean accuracies converged at lower values (head \(=0.193\), hand \(=0.177\), shoulder \(=0.149\)), and no differences were detected (\(H=0.33, p=0.849\)). Pairwise Mann–Whitney U tests with Holm correction confirmed no significant contrasts at any speed. Overall, the shoulder-mounted position yielded the highest average accuracy across most speeds, but these trends did not reach statistical significance.

\subsubsection{Overall Dynamic OCR Performance by Mounting Position}

\begin{figure}[ht]
\centering
\includegraphics[width=0.55\textwidth]{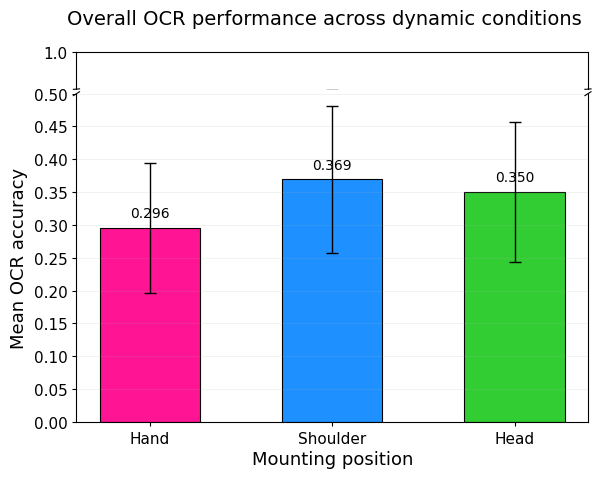}
\caption{
Overall dynamic OCR performance by mounting position. Bars show mean trial-level OCR accuracy pooled across all dynamic testing conditions, including walking speed, viewing angle, and camera type. Error bars indicate 95\% confidence intervals. The shoulder-mounted configuration achieved the highest mean accuracy, followed by the head-mounted and handheld configurations, although differences were not statistically significant. The y-axis includes a scale break above 0.5 to emphasize differences within the observed accuracy range while preserving the full scale up to 1.0.
}
\label{fig:ocr_position_overall}
\end{figure}

To provide a global comparison across mounting positions, we pooled trial-level OCR accuracies across all dynamic conditions and summarized them by mounting position (Figure \ref{fig:ocr_position_overall}). The shoulder-mounted configuration showed the highest mean trial-level OCR accuracy (Mean = 0.369, SD = 0.282), followed by the head-mounted configuration (Mean = 0.350, SD = 0.268) and the handheld configuration (Mean = 0.296, SD = 0.247).

However, the differences among mounting positions were not statistically significant (Kruskal--Wallis \(H = 0.89, p = 0.640\)). Pairwise Mann--Whitney U tests with Bonferroni correction also showed no significant contrasts between hand and shoulder, hand and head, or shoulder and head (all adjusted \(p = 1.0\)). Thus, although shoulder mounting yielded the highest overall mean OCR accuracy across dynamic trials, the advantage did not reach statistical significance.

\subsubsection{Accuracy vs Camera Type}

\begin{figure}[ht]
\centering
\includegraphics[width=0.5\textwidth]{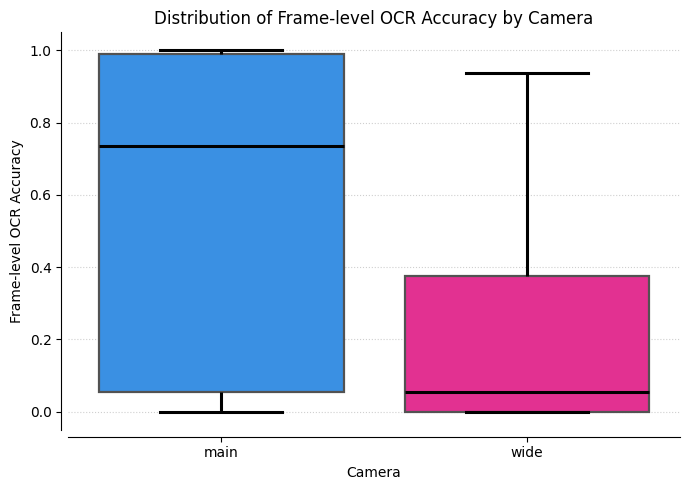}
\caption{
Distribution of frame-level OCR accuracy by camera type. The main camera achieved significantly higher accuracy (Mean = 0.575, Var = 0.173) compared with the ultra-wide camera (Mean = 0.229, Var = 0.097). Non-parametric statistical tests confirmed that the main camera consistently outperformed the ultra-wide camera.
}
\label{fig:ocr_camera}
\end{figure}

We compared frame-level OCR accuracy between the main and ultra-wide cameras (Figure~\ref{fig:ocr_camera}). The main camera achieved substantially higher accuracy (\(\text{Mean} = 0.575, \text{Var} = 0.173, N = 5690\)) than the ultra-wide camera (\(\text{Mean} = 0.229, \text{Var} = 0.097, N = 6090\)). Distribution plots showed a higher median and a shorter lower tail for the main camera, whereas the ultra-wide camera produced consistently lower values. Statistical tests corroborated this difference: a Kruskal--Wallis test (\(H = 1720.05, p < 0.001\)) and a Mann--Whitney U test (\(U = 2.49 \times 10^{7}, p < 0.001\)) both indicated significantly greater OCR accuracy for the main camera. These results highlight that camera type strongly influences OCR performance, with the main camera providing more reliable recognition than the ultra-wide camera.

\subsubsection{Meta Glasses Accuracy}

Meta glasses followed the same qualitative patterns as the smartphone cameras but with substantially lower accuracy under matched conditions. Across all conditions, the mean accuracy was 0.144 for Meta glasses, 0.494 for the iPhone 15 Pro Max main camera, and 0.183 for the iPhone ultra-wide camera, so Meta reached about 29\% of the iPhone main overall. 

\begin{figure}[ht]
\centering
\includegraphics[width=0.95\textwidth]{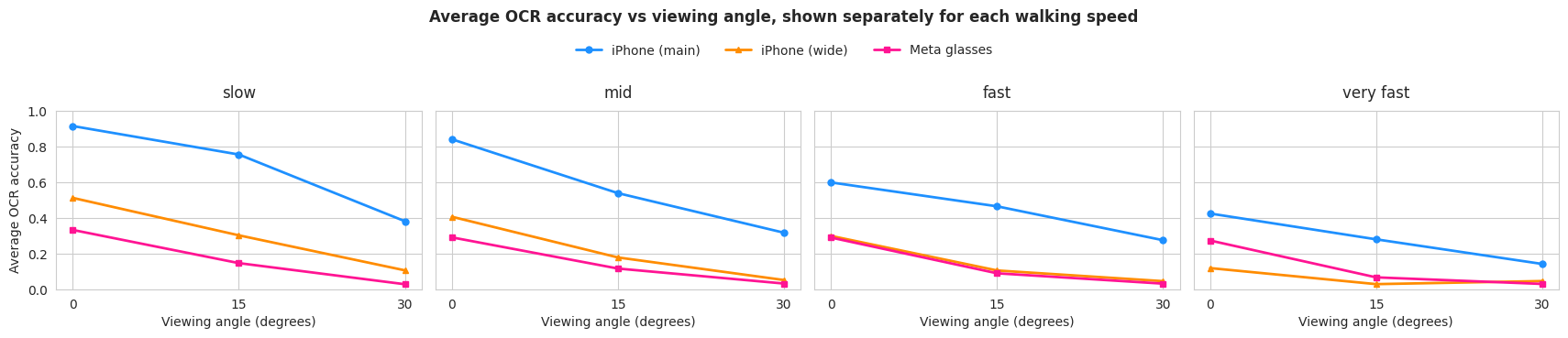}
\caption{
Average OCR accuracy versus viewing angle, shown separately for each walking speed. Lines depict iPhone 15 Pro Max (main), iPhone ultra-wide, and Meta glasses. Accuracy decreases with angle for all devices; Meta remains markedly lower and approaches zero at \(30^{\circ}\). All panels share the same \(y\)-axis.
}
\label{fig:meta_angle}
\end{figure}

Figure~\ref{fig:meta_angle} shows accuracy as a function of viewing angle for each walking speed. The Meta line is consistently well below both iPhone lines at every angle, with 0.296 at \(0^{\circ}\), 0.104 at \(15^{\circ}\), and 0.030 at \(30^{\circ}\) compared with 0.694, 0.509, and 0.278 for iPhone main, and 0.334, 0.153, and 0.062 for iPhone ultra-wide. Thus, Meta achieved about 43\% of iPhone main at \(0^{\circ}\), 20\% at \(15^{\circ}\), and 11\% at \(30^{\circ}\).

\begin{figure}[ht]
\centering
\includegraphics[width=0.95\textwidth]{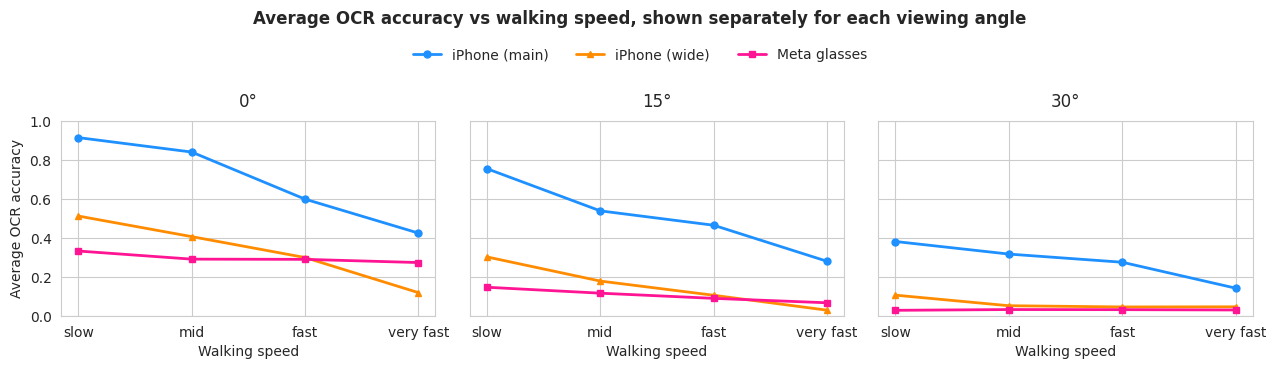}
\caption{
Average OCR accuracy versus walking speed, shown separately for each viewing angle. Lines depict iPhone main camera, iPhone ultra-wide camera, and Meta glasses. The iPhone main declines with speed. Meta has substantially low performance across all speeds. All panels share the same \(y\)-axis.
}
\label{fig:meta_speed}
\end{figure}

Figure~\ref{fig:meta_speed} shows accuracy as a function of walking speed at each viewing angle. For Meta, the speed pattern was shallow: 0.169 (slow), 0.146 (medium), 0.137 (fast), 0.123 (very fast). These differences were not statistically significant (Kruskal--Wallis \(H=0.54,\, p=0.91\)), and pairwise Mann--Whitney \(U\) tests with Holm correction confirmed no reliable contrasts. In contrast, iPhone main declined more strongly with speed over the same conditions (0.683 to 0.282). 

Overall, Meta performed much worse than the phone cameras. Although its mean accuracy is close to iPhone ultra-wide (0.144 vs 0.183), Meta’s field of view is narrower than the iPhone ultra-wide (99° vs 120°), which limits how often the sign stays within frame, especially at oblique angles.

\section{Discussion}

Our results demonstrate clear patterns in OCR performance across engines, camera positions, and fields of view. In all static testing scenarios, Google Vision achieved the highest accuracy overall, with PaddleOCR close behind. EasyOCR delivered moderate performance, while Tesseract performed poorly. Camera placement also influenced outcomes: the shoulder-mounted position yielded the highest average accuracy, although differences among positions did not reach statistical significance. Device type further shaped performance, as Meta glasses produced substantially lower accuracy compared with iPhone cameras. Finally, field of view introduced a pronounced trade-off. The main iPhone camera achieved higher recognition accuracy than the ultra-wide camera, though the latter maintained visibility of signs at wider angles. Taken together, these findings highlight the technical and practical factors that govern OCR reliability in assistive applications. They also underscore the tradeoffs across accuracy, device positioning, hardware quality, and environmental coverage that must be considered when designing OCR-based assistive technologies.

\subsection{Best Camera Position}

The shoulder-mounted position provided the highest overall mean OCR accuracy in the dynamic analysis. This pattern was observed both in the speed-specific comparisons and in the pooled summary across all dynamic conditions. Although this advantage was not statistically significant, the same descriptive trend appeared consistently, with shoulder outperforming head and hand on average in the pooled dynamic results. In contrast, handheld cameras are prone to misalignment as the hand tires and deviates from the intended direction, while head-mounted cameras introduce additional strain on the neck and can easily be misaligned with body orientation \cite{han2024wearables,faust2024technology}.
Moreover, head-mounted and handheld cameras may interfere with learned orientation and mobility techniques or restrict natural head movements that assist with auditory localization \cite{han2024wearables,faust2024technology}.

Although securing the camera to a backpack strap requires extra setup and is less intuitive than simply holding the device, the benefits are clear. Once properly mounted, the camera remains stable, freeing the user’s hands and minimizing motion artifacts caused by arm swing or grip fatigue. This stability is especially valuable during longer walking sessions, where handheld recording is difficult to sustain. Moreover, the fixed shoulder alignment ensures that the captured video better reflects the user’s natural heading, which simplifies downstream processing and interpretation.

An additional factor is the relative stability of the torso compared with other body segments. Prior work in gait analysis and wearable design has shown that torso motion during walking exhibits smaller oscillations and lower angular velocity than the head or hands, making it a more stable reference frame for wearable sensing \cite{han2024wearables,faust2024technology}. This inherent biomechanical stability has also made the torso a suitable mounting location for a wide range of electronic travel and orientation aids \cite{yang2024evaluating,feng2025haptics,ruan2025multi}. While comparisons between handheld and head-mounted cameras have shown similar functional outcomes for daily indoor tasks \cite{seiple2022comparison}, handheld systems may offer advantages for near-field text reading, whereas head-mounted devices can improve task time and head–body-movement efficiency—potentially reducing muscular strain \cite{seth2025does}. Currently, however, it remains unclear how torso-based wearables perform in these daily-use scenarios.

These findings align with our results, where the shoulder-mounted configuration yielded higher and more consistent OCR accuracy. The improved stability of the torso likely contributed to reduced frame jitter and more consistent sign appearance across successive frames—key factors for recognition performance \cite{abdullah2021new}. It is also worth noting that the camera on Meta glasses delivered substantially lower performance compared with the iPhone cameras, reflecting hardware and image-quality differences across devices \cite{han2024wearables, faust2024technology}.

\subsection{Best OCR Engine}

Our results demonstrate that Google Vision consistently achieved the highest OCR accuracy across all tested conditions. This outcome is expected, as it is a commercial service supported by large-scale datasets and continuous model refinement.

However, the performance gap between Google Vision and PaddleOCR was relatively small. PaddleOCR delivered competitive accuracy while remaining fully open-source. Moreover, PaddleOCR can run entirely on local hardware, whereas Google Vision requires cloud-based infrastructure. This has practical implications: cloud services introduce recurring costs, latency, and connectivity requirements, while PaddleOCR provides a more economical solution with lower computational demands relative to its accuracy. Beyond cost and efficiency, PaddleOCR offers additional advantages for assistive technology. Because it is open-source, the model can be customized or fine-tuned to recognize domain-specific text such as transportation signage or restaurant menus. Local execution also enhances privacy by eliminating the need to transmit environmental images to the cloud, and it ensures robustness in settings where internet access is unavailable or unreliable, such as subway areas. These qualities make PaddleOCR not only a technically competitive engine but also one that aligns with the real-world constraints of accessibility applications.

Taken together, these findings indicate that while Google Vision remains the top performer overall, PaddleOCR is the strongest open-source alternative. Its combination of accuracy, efficiency, adaptability, privacy, and offline capability makes it well suited for assistive technology systems. PaddleOCR represents a practical and high-performing choice for real-time, on-device OCR in accessibility applications.

\subsection{Trade-off Between Field of View and Distance}

Ultra-wide cameras capture a broader field of view, which ensures that signs located at 45° and 60° remain visible in the scene even at 7 meters. The advantage of a wider field of view is clear: more surrounding information can be captured and monitored, potentially enhancing user safety in assistive technology applications. However, this comes at the cost of reduced OCR performance. Because the same sign appears visually smaller in wide angle frames than in main-camera frames, recognition accuracy decreases substantially. In our experiment, the main camera achieved higher accuracy (Mean = 0.575, Var = 0.173) compared with the ultra-wide camera (Mean = 0.229, Var = 0.097).

This reflects an inherent accuracy and coverage trade-off: wider fields of view improve environmental awareness but reduce text legibility. For assistive technologies, this means that a single fixed camera may not be sufficient. One practical strategy is to use both cameras in complementary roles. For example, relying on the ultra-wide camera to detect and track potential signs at the periphery, then switching to the main camera or digitally zooming in for OCR once a candidate sign is identified. Such dynamic camera switching or hybrid processing would allow systems to balance coverage and recognition performance depending on user needs: maximizing safety by ensuring that relevant signs are not missed, while still providing the detailed accuracy required for text interpretation. This design principle highlights the importance of adaptive sensing strategies in real-world assistive applications.

\subsection{Design Implications for OCR-Based Assistive Navigation}

Beyond evaluating OCR engines and camera configurations, the present findings highlight several broader considerations for the design of assistive navigation technologies that incorporate text recognition. In practical applications, OCR should not function as an isolated capability but rather as part of a system that integrates sensing, device placement, and user interaction to support real-time mobility.

A critical implication concerns sensing strategy. Assistive navigation systems must simultaneously capture sufficient environmental context while maintaining adequate visual resolution for reliable text recognition. These results suggest that adaptive sensing approaches may be beneficial. For example, a wide-angle camera could monitor the surrounding environment to detect potential signage, while a higher-resolution camera performs detailed OCR once relevant text is identified. Such multi-stage sensing pipelines may allow assistive systems to balance environmental awareness with recognition accuracy while reducing the likelihood that important information is missed.

Device configuration also plays a role in system reliability. Camera placement influences image stability during locomotion, which in turn affects recognition performance. Wearable configurations that remain aligned with the user’s direction of travel may provide more consistent image capture during walking than configurations that move independently from the body. Designing assistive devices around stable body reference frames may therefore improve robustness while maintaining usability in everyday navigation.

Deployment considerations further influence the design of OCR-based assistive technologies. Although cloud-based OCR services often achieve high recognition accuracy, they introduce dependencies on network connectivity and may increase latency during continuous image processing. In contrast, open-source models capable of running locally on mobile devices provide advantages in terms of responsiveness, privacy, and reliability in environments where internet access is limited or unavailable. For assistive navigation systems that must operate continuously during travel, efficient on-device processing therefore represents an important practical consideration.

Importantly, OCR systems should be viewed as complementary tools within broader orientation and mobility (O\&M) strategies rather than replacements for existing techniques. Individuals with blindness or low vision commonly rely on mobility aids such as long canes or guide dogs, as well as auditory cues from the environment, to maintain spatial awareness. OCR can augment these strategies by providing access to textual information embedded in the environment, such as directional signage, room labels, or transportation information. In this role, OCR functions as an informational layer that supplements spatial awareness obtained through traditional mobility methods.

Many visually impaired travelers also rely heavily on auditory perception during navigation. Environmental sounds, reflections, and in some cases active echolocation are used to infer spatial structure and detect obstacles. Assistive technologies that incorporate OCR may therefore benefit from multimodal interaction designs that integrate textual information with auditory or tactile feedback. For example, recognized text could be delivered through context-aware audio cues while allowing environmental sounds to remain audible. Such multimodal systems may help users interpret textual information without disrupting existing navigation strategies.

Finally, recognition accuracy alone is not sufficient for effective assistive navigation. Systems must also determine when detected text is relevant to the user’s current task and deliver that information in a timely and accessible manner. Future research should therefore investigate how visually impaired users interpret OCR outputs during real mobility scenarios, how recognition errors influence navigation decisions, and how cognitive load is affected when multiple information sources are presented simultaneously.

Taken together, these considerations suggest that future OCR-based assistive technologies will benefit from integrated system designs that combine adaptive sensing, stable wearable hardware configurations, efficient on-device recognition, and multimodal interaction. Aligning these technical capabilities with established mobility practices may enable assistive systems that enhance environmental awareness while remaining compatible with the navigation strategies used by people with visual impairments.

\subsection{Limitations}

\subsubsection{Environmental Conditions}

All experiments were conducted indoors under uniform and controlled lighting conditions. This design ensured consistency across trials and minimized confounding factors, allowing clearer comparisons among OCR engines, camera positions, and device types. However, these conditions do not fully represent the environments in which assistive technologies are typically deployed. In real-world navigation scenarios, users encounter dynamic lighting, glare from reflective surfaces, shadows, moving pedestrians, and outdoor weather conditions such as rain, fog, or bright sunlight. Such factors can reduce contrast between text and background and introduce occlusions, motion blur, specular highlights, and perspective distortions that degrade OCR performance. Because these environmental variations were not included in the present experiments, the reported results likely represent an upper bound of OCR accuracy rather than performance under everyday navigation conditions.

\subsubsection{Scope of Tested Signage}

The experimental setup relied on a single sign board with a fixed spatial arrangement of signs and a limited number of text examples. While this configuration enabled systematic testing across distances, viewing angles, and walking speeds, it does not capture the full diversity of real-world signage. In practice, signs may contain low-contrast text, stylized or decorative fonts, handwritten characters, weathered or damaged surfaces, multiple languages, or complex layouts with multiple overlapping elements. Many also incorporate graphical components such as logos, symbols, or pictograms that can complicate text detection and recognition. Because these variations were not represented in the present dataset, the findings may not generalize to all types of signage encountered in real environments.

\subsubsection{Human Factors and Participant Population}

Walking experiments were conducted with a single sighted volunteer. The primary objective of the study was to evaluate OCR system performance under controlled conditions rather than to examine human navigation behavior. The participant repeatedly walked along a predefined path at controlled speeds, using a metronome to maintain consistent pacing. This approach helped isolate the effects of system variables, such as camera placement, camera hardware, and walking speed, while minimizing variability introduced by individual movement patterns.

Nevertheless, the use of a sighted participant represents an important limitation. PBLV often receive O\&M training and may adopt navigation strategies that differ substantially from those of sighted individuals \cite{feng2026residual,williams2013pray}. Moreover, in real-world travel, many individuals rely on mobility aids such as a white cane or guide dog, which influence body posture, arm motion, and walking dynamics. These factors may affect how wearable or handheld cameras are positioned relative to environmental targets. For instance, handheld configurations may be difficult to align consistently with body orientation when the hand is simultaneously engaged in cane use or other tasks, while shoulder-mounted cameras may occasionally be occluded by arm movements or cane sweeping. Head-mounted cameras may also be influenced by natural head movements during auditory orientation, as the head often turns toward sound sources during sound localization tasks \cite{thurlow1967head}. These human factors could affect both the stability of captured video and the visibility of environmental text, and therefore may influence OCR performance in ways not captured in the present controlled setup.

\subsubsection{Camera Hardware and Device Generalization}

This study evaluated only two devices: the iPhone 15 Pro Max and the Meta smart glasses. Although these devices represent widely available smartphone and wearable platforms, they do not capture the full range of hardware configurations used in assistive technology systems. Camera sensors differ substantially in resolution, dynamic range, lens quality, stabilization mechanisms, and on-device processing capabilities. Moreover, assistive technologies may employ a variety of camera placements and device configurations, including body-worn cameras, chest-mounted devices, head-mounted systems, and other wearable platforms. These differences in hardware design and mounting configuration can influence image quality, camera stability, and the visibility of environmental text. Consequently, the results reported here may not fully reflect the range of OCR performance that end-users would experience with other mobile or wearable assistive devices.

\subsubsection{Dynamic OCR Model Evaluation}

Dynamic walking experiments were conducted using PaddleOCR only. This model was selected based on the static benchmarking results, which showed that PaddleOCR achieved accuracy comparable to Google Vision while remaining open-source and capable of running locally without cloud processing. While this choice enabled efficient processing of large numbers of video frames, it limits the ability to directly compare dynamic performance across multiple OCR engines. Different OCR systems may exhibit varying sensitivity to motion blur, viewing angle, image resolution, and other factors associated with camera movement. Therefore, the dynamic performance results reported here should be interpreted as representative of PaddleOCR under the tested conditions rather than a comprehensive comparison across all OCR models.

\subsection{Future Directions}

Future work should extend these experiments to more realistic navigation environments. Testing in outdoor settings with variable lighting, glare, shadows, and dynamic pedestrian activity would provide a more comprehensive evaluation of OCR robustness under everyday conditions. The signage dataset should also be expanded to include multilingual text, stylized fonts, low-contrast print, and degraded or partially occluded signs. Evaluating OCR performance in common navigation contexts—such as overhead subway signage, airport wayfinding systems, and restaurant menu boards—would further improve ecological validity by introducing realistic variations in viewing angle, distance, and background clutter.

Human factors also warrant further investigation. Future studies should include a larger and more diverse participant population, particularly individuals with blindness or low vision using established mobility strategies such as long canes or guide dogs. Examining how users naturally hold or mount devices during navigation may provide additional insight into the interaction between human movement and OCR performance.

Additional hardware platforms should be evaluated, including a broader range of smartphones, alternative wearable cameras, and emerging AI-enabled smart glasses. Expanding both the participant population and the range of devices studied will help guide the development of OCR-based assistive technologies that remain reliable across diverse real-world conditions.

Future research should also evaluate a wider range of OCR engines under dynamic navigation conditions. The present study focused on PaddleOCR for dynamic testing because it provided competitive performance while supporting efficient local execution. However, other OCR systems—including additional open-source models and commercial services—may exhibit different sensitivities to motion blur, viewing angle, image resolution, and device hardware. Systematic comparisons across a broader set of OCR engines would provide a more comprehensive understanding of algorithm robustness and help guide the selection of recognition models for assistive navigation systems.

\section{Conclusion}

This study demonstrates that OCR performance is shaped by multiple interacting factors, including walking speed, camera mounting position, lens type, and the choice of OCR engine. Among all engines tested, Google Vision, a commercial cloud-based service, provided the highest accuracy. However, PaddleOCR, an open-source engine, achieved competitive results and was the strongest non-commercial option, underscoring its potential for real-time, on-device applications.

Walking speed had a significant impact on performance, with accuracy declining as speed increased. Camera position did not significantly affect OCR accuracy; however, the shoulder-mounted configuration had the highest average accuracy. The ultra-wide camera provided a broader field of view, but this advantage came at the cost of reduced OCR accuracy.

These findings emphasize the importance of context-aware design in assistive technology. OCR engine selection cannot be made in isolation; system designers must also account for mobility dynamics (e.g., walking speed and head motion), camera placement (head, shoulder, or hand), and device characteristics (main vs. ultra-wide lens). Optimizing exclusively for field of view may reduce recognition accuracy, while restricting field of view risks missing critical contextual information. Careful trade-offs are therefore necessary to balance safety, coverage, and text-recognition performance.

Looking forward, integrating robust open-source OCR engines such as PaddleOCR into wearable or handheld platforms offers a pathway toward affordable, real-time solutions that run locally, without reliance on cloud connectivity. To fully support individuals with vision loss, OCR-based assistive technologies must be validated across diverse environments, signage types, and user populations. Only through such comprehensive evaluation can these systems become reliable tools for daily navigation and independent mobility.

\section{Acknowledgment}
This research was supported by the National Science Foundation under Grant Nos. ITE-2345139, ECCS-1928614, CNS-1952180,
and ITE-2236097 by the National Eye Institute and Fogarty International Center under Grant No. R21EY033689, as
well as by the U.S. Department of Defense under Grant No. VR200130. The content is solely the responsibility of the authors and does not necessarily represent the official views of the National Institutes of Health, National Science Foundation,
and Department of Defense.

This study used ChatGPT-5 to aid in generating Python scripts for data analysis and for refining grammar and language throughout the manuscript. The authors carefully reviewed and verified all model outputs to ensure accuracy and uphold the intellectual integrity of the research.

\section{Disclosure of interest}

The authors report no conflicts of interest. 

\vspace*{ 1 cm}

\bibliographystyle{ieeetr}
\bibliography{references}

\end{document}